\documentclass{ngsm2024} 

\usepackage{booktabs}
\usepackage{amsmath}
\usepackage{adjustbox}

\title[Q-S5: Towards Quantized State Space Models]{Q-S5: Towards Quantized State Space Models}

\optauthor{
\Name{Steven Abreu$^*$} \Email{s.abreu@rug.nl}\\
\addr University of Groningen \& Intel Labs
\AND
\Name{Jens E. Pedersen$^*$} \Email{jeped@kth.se}\\
\addr KTH Royal Institute of Technology
\AND
\Name{Kade M. Heckel$^*$} \Email{kmh70@cam.ac.uk}\\
\addr University of Cambridge
\AND
\Name{Alessandro Pierro} \Email{alessandro.pierro@intel.com}\\
\addr LMU Munich \& Intel Labs
\begin{center}
\end{center}
}

\newcommand{\textul}[1]{\underline{#1}}
\newcommand{\textrd}[1]{#1}

\begin{document}

\maketitle

\vspace{-1.5cm}
\begin{abstract}
In the quest for next-generation sequence modeling architectures, 
State Space Models (SSMs) have emerged as a potent alternative to transformers, particularly for their computational efficiency and suitability for dynamical systems.
This paper investigates the effect of quantization on the S5 model to understand its impact on model performance and to facilitate its deployment to edge and resource-constrained platforms.
Using quantization-aware training (QAT) and post-training quantization (PTQ), we systematically evaluate the quantization sensitivity of SSMs across different tasks like dynamical systems modeling, Sequential MNIST (sMNIST) and most of the Long Range Arena (LRA). 
We present fully quantized S5 models whose test accuracy drops less than 1\% on sMNIST and most of the LRA. 
We find that performance on most tasks degrades significantly for recurrent weights below 8-bit precision, but that other components can be compressed further without significant loss of performance.
Our results further show that PTQ only performs well on language-based LRA tasks whereas all others require QAT. 
Our investigation provides necessary insights for the continued development of efficient and hardware-optimized SSMs.
\end{abstract}



\section{Introduction}

Sequence modeling architectures based on state space models (SSMs) \cite{voelker_legendre_2019,gu_hippo_2020,gu_efficiently_2022,smith2023simplified} have emerged as a powerful sequence modeling framework. SSMs scale better than transformers with sequence length and show state-of-the-art performance on many sequence modeling tasks \cite{smith2023simplified,orvieto_resurrecting_2023}. 
While many prior works have investigated quantization \cite{hubara_quantized_2018,guo_survey_2018,gholami_survey_2021} to reduce the computational cost of transformer architectures \cite{zhou_survey_2024,wang_beyond_2024,tang_survey_2024,ma_era_2024}, 
the effect of quantization on SSMs is not widely studied.

It is well-known that quantizing RNNs comes with more challenges than feed-forward networks \cite{ott_recurrent_2017,li_quantization_2021}, thus it is not clear how SSM architectures perform under quantization constraints. However, low-precision weights and activations for recurrent neural networks (RNNs) are widely used in neuromorphic computing \cite{bos2023sub}, yielding models with SOTA performance and energy efficiency \cite{shrestha_efficient_2024}.
The earliest SSM model, the LMU \cite{voelker_legendre_2019}, presented a state-of-the-art spiking network model (\textit{i.e.}, using 1-bit activations) and later achieved state-of-the-art accuracy and energy efficiency when quantized for efficient hardware implementation \cite{blouw2021hardware,Gaurav2022spiking}. 

To the best of our knowledge, subsequent SSM models based on HiPPO \cite{gu_hippo_2020,gu_efficiently_2022,orvieto_resurrecting_2023,smith2023simplified,gupta_diagonal_2022,gu_parameterization_2022} have not been quantized before.
%
%
In this work, we examine the S5 model \cite{smith2023simplified} and conduct experiments using quantization-aware training (QAT), where the model is trained with dynamic quantization, and post-training quantization (PTQ), where a full-precision model is quantized without training. 

We provide the first step towards efficient quantized SSMs by (1) examining the relationship between performance and the parameter precision of different SSM components when evaluated on both a dynamical system and on the Long Range Arena (LRA), and (2) presenting the first fully quantized SSMs that performs well on most of the LRA.

\begin{figure}[t]
    \centering
    \includegraphics[width=\textwidth]{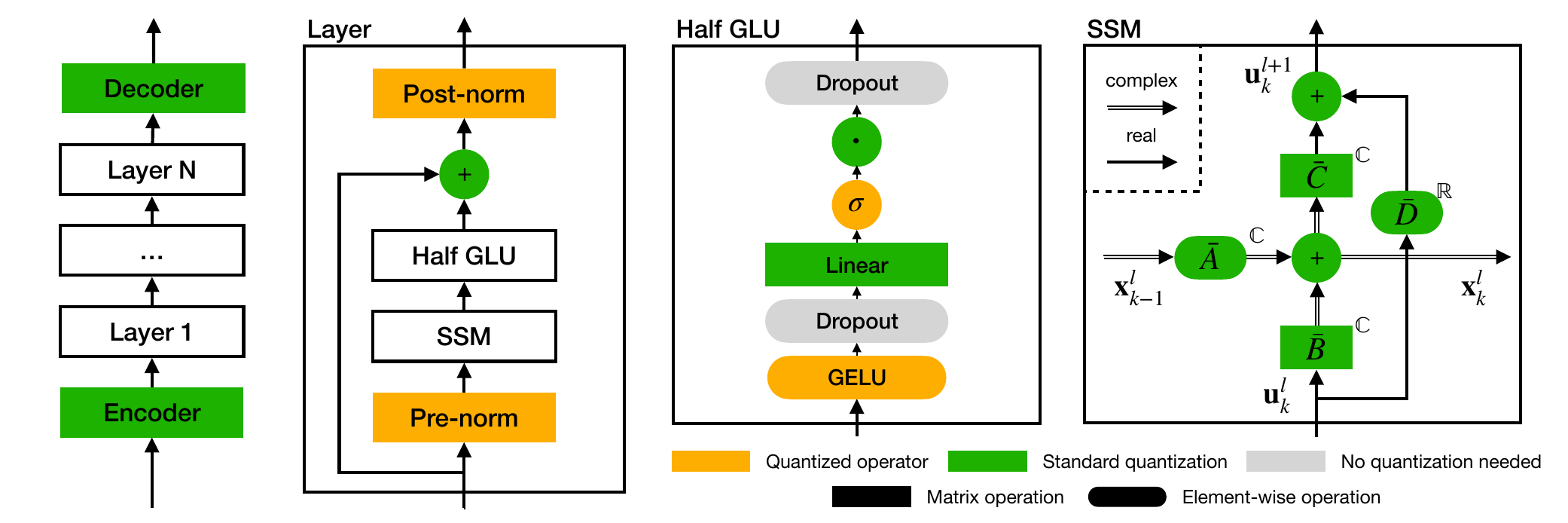}
    \caption{
    The model architecture used in this paper, based on S5 \cite{smith2023simplified}. 
    }
    \label{fig:architecture}
\end{figure}

\section{Methods}

\paragraph{S5 model}
We use the S5 architecture \cite{smith2023simplified} because of its simplicity and focus on the core aspects of SSM functionality. The discretized dynamics of the S5 model are: 
\begin{align}
  x_k &= \bar{A}x_{k-1} + \bar{B}u_k \\
  y_k &= \bar{C}x_{k-1} + \bar{D}u_k
\end{align}
where $\bar{A} \in \mathbb{C}^P$ is the diagonal recurrent matrix, $\bar{B} \in \mathbb{C}^{P \times H}$ is the input matrix, $\bar{C} \in \mathbb{C}^{H \times P}$ is the output matrix, $\bar{D} \in \mathbb{R}^{H \times H}$ is the skip connection matrix, $u_k \in \mathbb{R}^{H}$ is the input to the S5 block at timestep $k$, $x_k \in \mathbb{R}^P$ is the S5 block's hidden state at time step $k$, and $y_k \in \mathbb{R}^H$ is the S5 block's output at timestep $k$. 


\paragraph{Quantization}
We use the accurate quantized training (AQT) library in JAX for our quantization experiments, which has been used in previous work on quantization-aware training \cite{zhang_binarized_2023,zhang_pokebnn_2022,abdolrashidi_pareto-optimal_2021,ding_4-bit_2023}. We denote the tensor to be quantized with $\mathbf{x}$ and the number of bits to use with $n$, then the quantized tensor $\mathbf{x}_n$ is defined as:
\begin{align}
    \bar{\mathbf{x}}_{n} = \left\lfloor \frac{(2^{n-1}-1) \mathbf{x}}{\max | \mathbf{x} |} \right\rceil = \left\lfloor \frac{\mathbf{x}}{\Delta_x} \right\rceil = \left\lfloor s_x \mathbf{x}\right\rceil
\end{align}
where $\lfloor \cdot \rceil$ indicates rounding to the nearest integer and $s_x = (2^{n-1}-1) (\max |\mathbf{x}|)^{-1}$ is the scale for the given tensor $\mathbf{x}$, and $\Delta_x$ is the corresponding step size.  
As highlighted in Figure \ref{fig:architecture} by the orange boxes, we replace the normalization operation and the GELU activation function with quantized variations thereof. Further details on our quantization implementation are in Appendix \ref{appendix:quantization-details}.

\section{Experiments}

To investigate how different components of the S5 architecture are affected by quantization errors, we vary the level of quantization separately for the different components shown in Figure \ref{fig:architecture}. We distinguish between the quantization of weights ($W$) and activations ($A$), and further distinguish between the parameter precision of SSM components and non-SSM components. 
We begin by studying the effect of quantization in a simple dynamical setting before moving on to sequential MNIST and Long Range Arena tasks.

\subsection{Prediction of dynamical systems with QAT}

\begin{figure}[t]
    \centering
    \includegraphics[width=\linewidth]{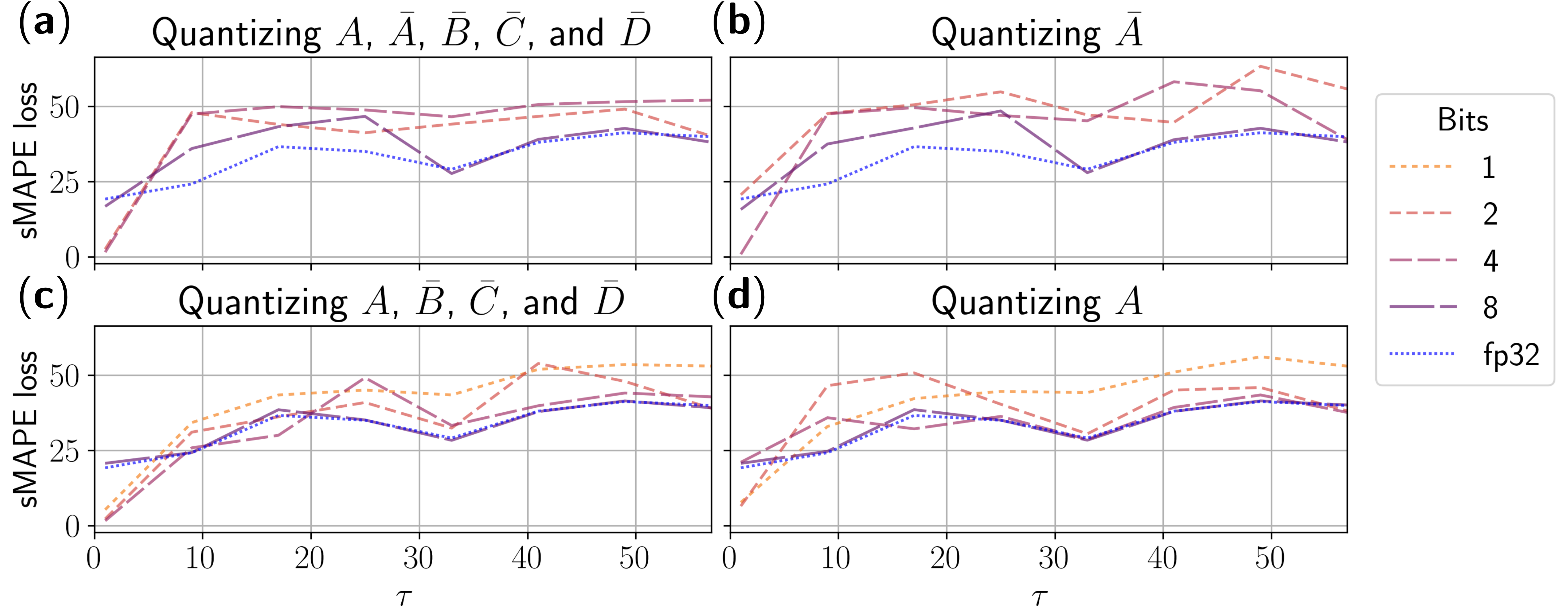}
    \caption{
    sMAPE loss for the Mackey-Glass dataset for temporal delays $\tau \in [0, 100]$ under 4 quantization settings, averaged across 4 runs.
    In (\textbf{a}) and (\textbf{b}), $\bar{A}$ at 1 bit precision never converged and is omitted.
    }
    \label{fig:dynamical}
\end{figure}

We study a time-delayed 10-dimensional Mackey-Glass system (see Section \ref{app:mackey_glass}) under varying quantization configurations for a small two-layer S5 network (see Figure \ref{fig:architecture}) with a total of 624 trainable parameters, using QAT.
Since $\bar{A}$ linearly maps the recurrent dynamics $x_{k-1}$, Figure \ref{fig:dynamical} (\textbf{b}) shows that the quantization of $\bar{A}$ significantly degrades the model's ability to capture temporal relationships.

Quantizing the activation should similarly degrade performance, since it truncates the amount of information between the layers.
Panel \textbf{d} supports this, although only in the extreme case of single-bit activations.
However, when we additionally quantize the $\bar{B}$, $\bar{C}$, and $\bar{D}$ matrices in panel \textbf{c}, performance worsens slightly, although with a more graceful degradation in time.

\subsection{Sequential MNIST and Long Range Arena}

Given our insights above, we now move on to report results on the sequential MNIST (sMNIST) task and on all Long Range Arena (LRA) tasks \cite{tay_long_2020} except Path-X. 
In addition to QAT, we also report results for PTQ where the full-precision network is quantized without any re-training. 
We further present promising preliminary results on quantization-aware fine-tuning on sMNIST in Appendix \ref{appendix:qaft}, as well as additional ablation studies on the quantization of activation functions and layer normalization in Appendix \ref{appendix:ablations-smnist-quantized-ops}. All details on the experiments are in Appendix \ref{appendix:smnist-lra-experiment-details}.

\begin{table}[t]
\caption{Test accuracy (in \%) for sMNIST and LRA tasks for QAT, PTQ and full-precision. The best quantized result is \textbf{bold}, all results $<$ 10 percentage points decrease are \textul{underlined}.}
\begin{adjustbox}{center}
\begin{tabular}{llcccccc}
\toprule
    & & \textbf{SMNIST} & \textbf{ListOps} & \textbf{Text} & \textbf{Retrieval} & \textbf{sCIFAR} & \textbf{Pathfinder} \\
\midrule
    & FP \cite{smith2023simplified}  & 99.65 & 62.15 & 89.31 & 91.4 & 88 & 95.33 \\
\midrule
PTQ & W8A8          & 96.27          & 26.65          & \textbf{88.49} & \textul{89.87} & 44.83          & 50.90          \\
\midrule
QAT & W8A8          & \textul{99.54} & \textbf{39.05} & 57.39          & \textul{86.26} & \textul{86.95 }& 50.81          \\
    & W4A8SSM8      & \textbf{99.63} & 36.80          & 50.72          & \textul{82.78} & \textbf{87.20} & \textbf{95.06} \\
    & W4A8\={A}8    & \textul{99.26} & 37.15          & \textul{87.79} & \textbf{90.81} & \textul{82.56} & 53.21          \\
    & W4A8          & \textrd{12.68} & 37.35          & 50.00          & 49.39          & \textrd{10.00} & 50.54          \\
    & W2A8SSM8      & \textul{99.56} & 36.80          & 52.21          & 72.15          & \textul{85.57} & \textul{94.34} \\
    & W2A8\={A}8    & 80.76          & 36.70          & \textul{81.26} & 73.92          & 37.02          & 52.63          \\
    & W2A8          & 54.75          & 23.15          & 74.21          & 79.70          & 31.01          & 50.70          \\
\bottomrule
\end{tabular}
\end{adjustbox}
\label{tbl:quant-results}
\end{table}

The results on LRA reported in \autoref{tbl:quant-results} suggest various insights. Firstly, almost all configurations using less than 8 bits for $\bar{A}$ results in poor task performance. This mirrors findings for quantization of the LMU model \cite{blouw2021hardware}.
Secondly, as observed in previous work on quantization of attention-based LLMs \cite{yin2024junk}, the relationship between quantization precision and task performance is non-linear. We hypothesize that this can be caused by the deep double descent phenomenon, with the quantization pushing the initially over-parameterized model into the critical region \cite{Nakkiran_2021}.
Extensive hyperparameter optimization and additional experiments may determine better QAT recipes.

Notably, no quantized model was able to come close to the performance of the full-precision model on ListOps. However, the accuracy still exceeds all baselines from the LRA paper \cite{tay_long_2020} which shows that the model is indeed learning. It is also the only task where a model with quantized recurrent weights performs similarly to the best model.

\section{Outlook} 

We presented the first fully quantized SSM models based on the S5 architecture. Our quantized models learn all but one of the LRA tasks to within 1 percentage point of the full-precision model's accuracy while using $>$4x less memory and almost exclusively integer operations.
We further showed that PTQ is competitive with QAT for language-based LRA tasks (Text and Retrieval) and included promising preliminary results on quantization-aware finetuning (QAFT) in Appendix \ref{appendix:qaft}.

In the future, we hope to expand our QAFT methods to explore optimal tradeoffs between training compute and final model efficiency. We further hope that better QAFT methods will expand our methods to large pre-trained selective SSMs like Mamba \cite{gu2024mamba}, Jamba \cite{lieber2024jamba}, and Griffin \cite{de2024griffin}.

Theoretically, we hope to extend our analysis to demonstrate
how sparse binary activations (spikes) can code for complex patterns in spatio-temporal signals using work based on linear kernels \cite{Pedersen_Conradt_Lindeberg_2024}, by approximating spatial and temporal dynamics with recurrent linear maps, similar to the approach taken in SSMs \cite{voelker_legendre_2019,gu_hippo_2020}. 

\newpage
\clearpage
\bibliography{sample}


\newpage
\clearpage
\appendix

\section{Implementation details on quantization}
\label{appendix:quantization-details}

We use dynamic quantization with symmetric per-tensor scales, such that the zero point is preserved, that is, the mean remains unchanged. The scale is computed separately for each batch. As such, there is only one scale per weight matrix, and one scale per activation vector.

\subsection{Quantization-aware training}

During the backward pass, we use straight-through estimators for the rounding operations, \textit{i.e.}, $\frac{d}{dx} \lfloor x \rceil = 1$.
For simplicity, we quantize the forward and backward computations during training with equal precision in all experiments. 
The summation results of dot product operations are accumulated in 32-bit integers to prevent overflow.

\subsection{Quantized normalization}

We quantize the layer normalization operation \cite{li_quantization_2021} and we do not use batch normalization in our quantized models because it is incompatible with the typical single/low-batch mode of inference that is commonly used in resource-constrained environments.

\subsection{Quantized GELU: qGELU}
\label{appendix:qgelu}

\begin{figure}[h]
    \centering
    \includegraphics[width=0.7\linewidth]{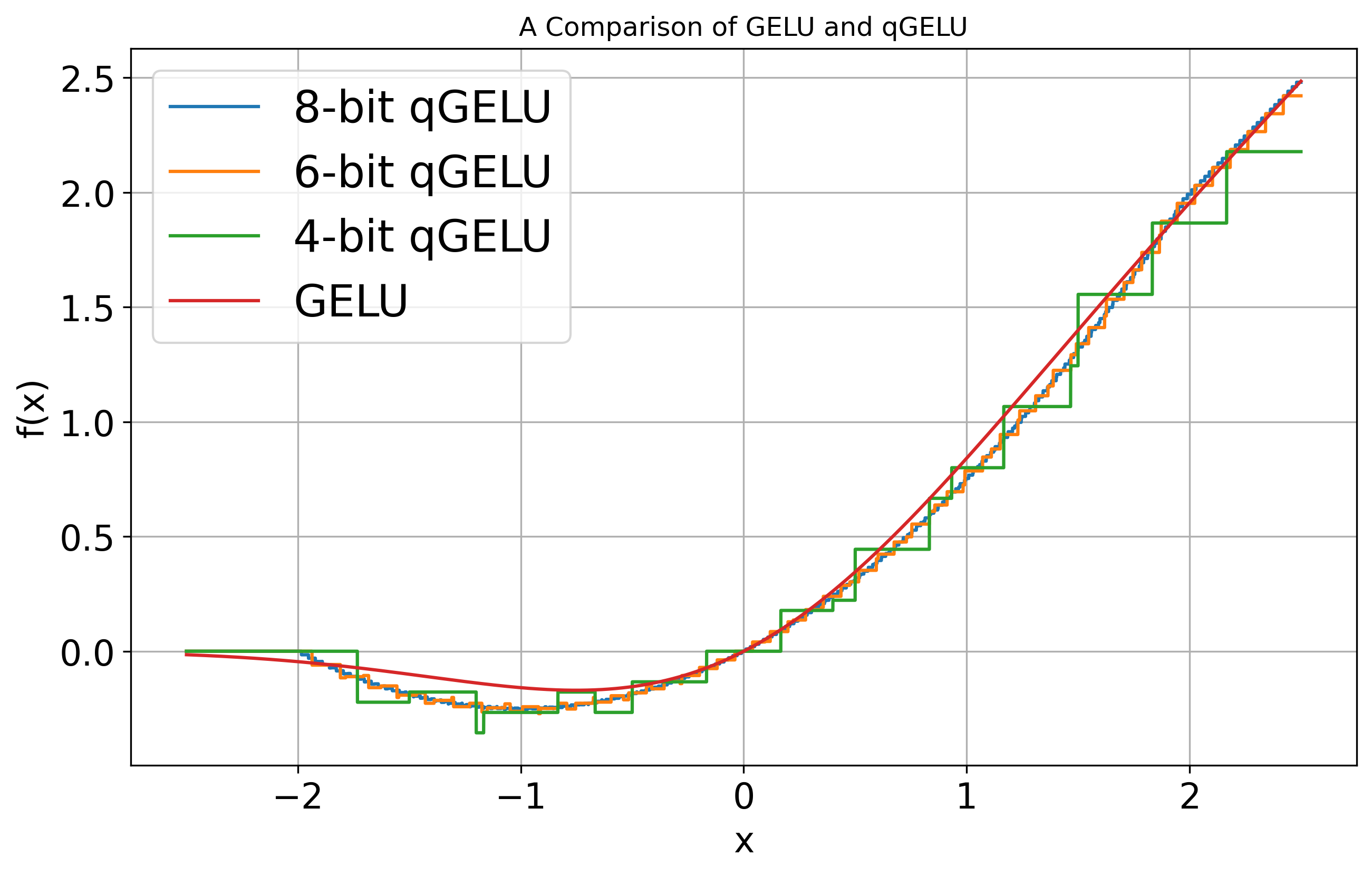}
    \caption{qGELU activation function operating on differing levels of quantized inputs. Note that below 8 bit quantization, the qGELU function begins to systematically underestimate the output of GELU.}
    \label{fig:qgelu}
\end{figure}

For quantizing the activation function used in the S5 architecture, we take an approach similar to other works in quantized computer vision research. MobileNetV3 \cite{mobilenet} uses a hard-swish variant which uses the approximation $swish(x) = x\sigma(x) \approx x\cdot ReLU6(x+3)/6 = hard\_swish(x)$, where $ReLU[n](x) = max(min(x+shift, n), 0)$.  These shifted and bounded $ReLU[n]$ operations have hardware supported SIMD operations on NVIDIA hardware. We use a different parametrization based around $ReLU4$ because it both closely matches the GELU\cite{hendrycks2023gaussian} function used in the S5 architecture. This variant, which we term qGELU, can be implemented for integers using a bit shift operation instead of division, making it more efficient.

Figure \ref{fig:qgelu} illustrates varying quantization levels of the qGELU function used in this work. Notably, 6 and 4 bit qGELU functions begin to underestimate the activation value appreciably in the saturation regime ($x > 2$), while 8-bit qGELU remains quite close. Currently, NVIDIA hardware only supports SIMD $ReLU[n]$ operations for 16-bit signed integers at the smallest, meaning that using 8 bit activations will not yield greater computational speedup during training. As lower precision data types become cemented within deep learning, it would be exciting if future GPU architectures will support vectorized $ReLU[n]$ operations on signed 8-bit integers. By using $ReLU4$, qGELU has the distinct advantage that its hard-sigmoid function can be implemented using the right bit-shift operator, avoiding the need for division; this has great implications for the design of ASICs for inference computation, as the circuitry for computing layer activations can be dramatically simplified.

\section{Details for the experiments}

\subsection{Mackey-Glass system} \label{app:mackey_glass}

\begin{figure}[h]
    \centering
    \includegraphics[width=1\linewidth]{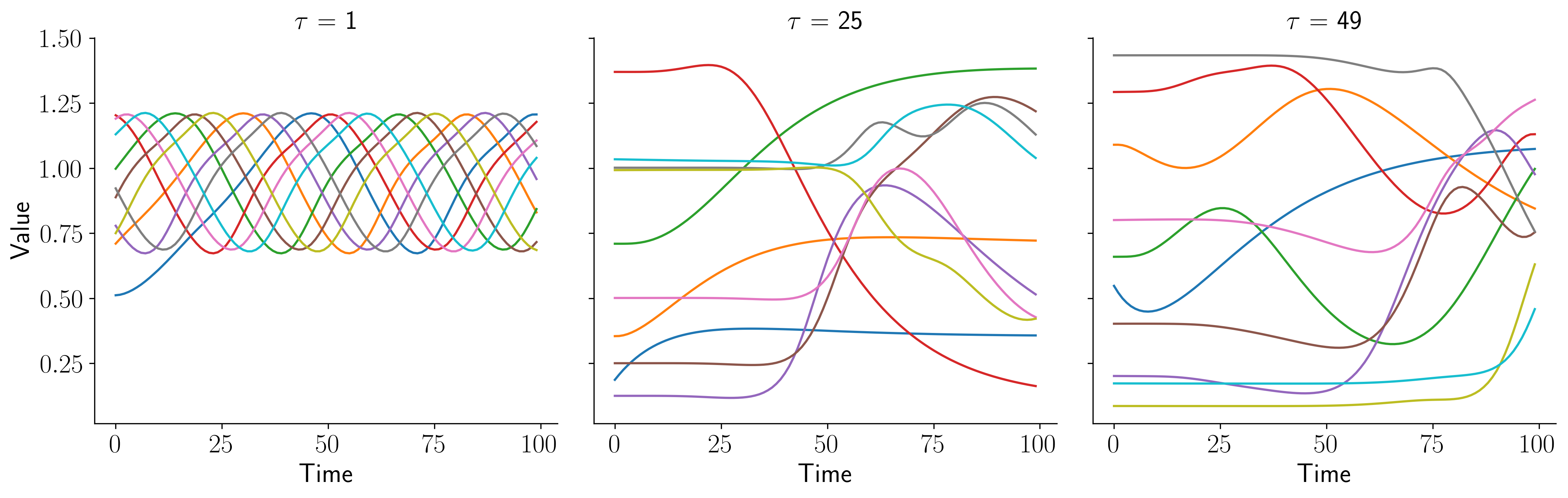}
    \caption{
    The first 100 timesteps of three samples from the 10-dimensional Mackey-Glass system under varying time-delay values $\tau$.
    }
    \label{fig:mackey-glass}
\end{figure}

The Mackey-Glass system studied in this work was generated using the Dysts\cite{gilpin2023model, gilpin2023chaos}  dynamical systems dataset generation library. We characterized a 10-dimensional Mackey-Glass system by the following equation, starting from a random initial condition for 1024 timesteps using standard forward-Euler integration (see Figure \ref{fig:mackey-glass}):
\begin{equation}
Q'(t) = \frac{\beta Q(t - \tau)}{1 + Q( t - \tau)^n } - \gamma Q(t)
\end{equation}
Belonging to the family of Delay Differential Equations, Mackey-Glass systems are an attractive benchmark system that have been utilized in previous works\cite{gu_hippo_2020} to study the memory capacity of recurrent model architectures. Specifically, their modifiable delay $\tau$ enables the study of a model's ability to retain information and capture long-term dependencies; by carrying out an ablation of computational precision and analyzing the model's prediction error versus $\tau$, insights into the design space of quantized SSMs can be extracted.

To characterize model performance on the Mackey-Glass prediction task, we adopt the Symmetric Mean Average Percent Error (sMAPE) by Dysts\cite{gilpin2023model} due to its relation with the rate of separation (via the Lyapunov exponent) and its popularity as a metric in dynamical systems forecasting:

$$
\epsilon(t) \equiv \frac{200}{t} \sum_{t'=1}^t \frac{|y(t') - \hat{y}(t')|}{|y(t')| + |\hat{y}(t')|}
$$

\subsection{sMNIST and LRA}
\label{appendix:smnist-lra-experiment-details}

For training the full-precision S5 networks on sMNIST and LRA, we use the exact same training setup and hyperparameters as the original S5 paper \cite{smith2023simplified}. 
For all QAT runs, we replace the normalization layer with our quantized layer normalization layer described in Appendix \ref{appendix:quantization-details}, the GELU activation function with our quantized GELU function described in Appendix \ref{appendix:qgelu} and the sigmoid function with the hard sigmoid function. Ablations on these replacements are presented in Appendix \ref{appendix:ablations-smnist-quantized-ops} on the sMNIST task. 
Aside from these changes to the model definition, we use the same training setup and hyperparameters as for the full-precision training, except for the Retrieval and Pathfinder tasks where we scaled down the original learning rate by $0.75$ to improve convergence.
We report the best test accuracies from the entire training run.

\section{Additional experimental results}

We present additional experimental results on the sMNIST task and the LRA tasks with more quantization configurations for PTQ and different full-precision (FP) models in Table \ref{fig:full-results-lra-smnist}. 

\begin{table}[h]
\centering
\caption{
    All results from our experiments on the sMNIST task and the LRA tasks, expanded from Table \ref{tbl:quant-results}.
    Results show test accuracies.
}
\begin{tabular}{l | l | c c c c c c}
\toprule
    & Model                                 & sMNIST & ListOps & Text   & Retrieval & sCIFAR & Pathfinder \\
    & (Input length)                        & (784)  & (2024)  & (4096) & (4000)    & (1024) & (1024)     \\
\midrule
    & S5 results \cite{smith2023simplified} & 99.65  & 62.15   & 89.31  & 91.4      & 88     & 95.33      \\
    & Our reproduction                      & 99.54  & 61.1    & 88.77  & 91.34     & 87.76  & 94.33      \\
    & - using layer norm                    & 99.54  & 57.9    & 88.74  & 90.78     & 85.51  & 85.96      \\
\midrule
PTQ & W8A8*                                 & 96.74  & 35.4    & 88.61  & 90.26     & 43.33  & 50.90      \\
    & W8A8                                  & 96.27  & 26.65   & 88.49  & 89.87     & 44.83  & 50.89      \\
    & W4A8SSM8                              & 95.99  & 31.45   & 88.2   & 88.67     & 44.71  & 50.87      \\
    & W4A8\={A}8                            & 37.98  & 9.65    & 88.07  & 87.35     & 14.35  & 51.07      \\
    & W4A8                                  & 9.74   & 17.8    & 87.64  & 77.26     & 9.99   & 49.61      \\
    & W2A8SSM8                              & 61.94  & 17      & 50.43  & 55.18     & 22.53  & 49.81      \\
    & W2A8\={A}8                            & 11.35  & 8.45    & 54     & 50.51     & 9.75   & 50.55      \\
    & W2A8                                  & 11.35  & 17.25   & 51.18  & 50.61     & 9.99   & 50.55      \\
\midrule
QAT & W8A8                                  & 99.54  & 39.05   & 57.39  & 86.26     & 86.95  & 50.81      \\
    & W4A8SSM8                              & 99.63  & 36.8    & 50.72  & 82.78     & 87.2   & 95.06      \\
    & W4A8\={A}8                            & 99.26  & 37.15   & 87.79  & 90.81     & 82.56  & 53.21      \\
    & W4A8                                  & 12.68  & 37.35   & 50.00  & 49.39     & 10.00  & 50.54      \\
    & W2A8SSM8                              & 99.56  & 36.8    & 52.21  & 72.15     & 85.57  & 94.34      \\
    & W2A8\={A}8                            & 80.76  & 36.7    & 81.26  & 73.92     & 37.02  & 52.63      \\
    & W2A8                                  & 54.75  & 23.15   & 74.21  & 79.70     & 31.01  & 50.70      \\
\bottomrule
\end{tabular}
\label{fig:full-results-lra-smnist}
\end{table}

\subsection{Ablations on quantized operators}
\label{appendix:ablations-smnist-quantized-ops}

We present ablation studies on the effect of our proposed quantized GELU activation function (see Appendix \ref{appendix:qgelu}), the hard sigmoid, and the quantized layer norm operation (see Appendix \ref{appendix:quantization-details}). 

Table \ref{fig:smnist-ablation-qgelu-qln} shows the test accuracy of sMNIST when the model is trained using QAT with different configurations of using GELU or qGELU and batch normalization (BN) or quantized layer normalization (qLN). 
The results show that our proposed quantized GELU activation function, used in conjunction with the quantized layer normalization does not lead to a significant dropoff in performance. On training runs that achieve at least 99\% test accuracy, the performance change from using qGELU and qLN is on average -0.04 percentage points (pp).

Table \ref{fig:smnist-ablation-ptq} further shows an ablation study on sMNIST using post-training quantization (PTQ) based on the W8A8 quantization. Results show that the use of the hard sigmoid activation function leads to the largest performance degradation -- 0.44 pp relative to the total drop of 0.47 pp using all three replacements. As such, we are motivated to investigate the effect of the hard sigmoid on quantization-aware training and fine-tuning and possibly find better substitutions for the sigmoid function in future work. 

\begin{table}[h]
\centering
\caption{
    Ablation study of the proposed quantized GELU activation function and the quantized layer norm. Results show the test accuracy on sMNIST. The rightmost column shows the absolute change in accuracy from using GELU\& BN to using qGELU \& qLN. \\
    $^*$) Baseline run is using full precision and therefore uses GeLU, not qGeLU.
}
\begin{tabular}{ l  c  c  c  c}
\toprule
 & GeLU \& BN & qGeLU \& BN & qGeLU \& qLN & Change (pp) \\
\midrule
FP              & \textbf{99.53\%}    & n/a                 & \textbf{99.48\%}$^*$      & -0.05 \\
W8A8            & \textbf{99.48\%}    & \textbf{99.47\%}    & \textbf{99.54\%}          & +0.06\\
W4A8 (SSM: W8)  & \textbf{99.52\%}    & \textbf{99.52\%}    & \textbf{99.63\%}          & +0.11\\
W4A8 ($\bar{A}$: W8)  & 97.32\%             & \underline{98.03\%} & \textbf{99.26\%}    & -0.06 \\
W4A8            & 9.80\%              & 34.51\%             & 12.68\%                   & (+2.88)\\
W2A8 (SSM: W8)  & \textbf{99.64\%}    & \textbf{99.45\%}    & \textbf{99.56\%}          & -0.08\\
W2A8 ($\bar{A}$: W8)  & 71.63\%             & 82.40\%             & 80.76\%             & (+9.13)\\
W2A8            & 46.07\%             & 46.77\%             & 54.75\%                   & (+8.68)\\
\midrule
W8A4 (SSM: A8) & - & - & 95.63\% & - \\
W8A4           & - & - & 78.23\% & - \\
W8A2 (SSM: A8) & - & - & 25.47\% & - \\
W8A2           & - & - & 18.87\% & - \\
\bottomrule
\end{tabular}
\label{fig:smnist-ablation-qgelu-qln}
\end{table}

\begin{table}[h]
\centering
\caption{Ablation study of the proposed quantized GELU activation function, the hard sigmoid replacement for the sigmoid function, and the quantized layer norm. The baseline result shows the test accuracy of a full precision network trained with layer normalization on sMNIST.}
\begin{tabular}{lccccc}
\toprule
 & HS & qGELU & qLN & Test accuracy & Change (pp) \\
\midrule
Baseline &  &  &  & 99.54\%                                 & - \\
W8A8 LN &  &  &  & 96.74\%                                  & 0.0\\
W8A8 LN & \checkmark &  &  & 96.30\%                        & -0.44 \\
W8A8 LN &  & \checkmark &  & 96.75\%                        & +0.01 \\
W8A8 LN &  &  & \checkmark & 96.79\%                        & +0.05 \\
W8A8 LN & \checkmark & \checkmark & \checkmark & 96.27\%    & -0.47 \\
\bottomrule
\end{tabular}
\label{fig:smnist-ablation-ptq}
\end{table}

\subsection{Quantization-aware fine-tuning}
\label{appendix:qaft}

We present results on quantization-aware fine-tuning (QAFT) where the floating-point baseline model was used to initialize the weights before doing QAT. As commonly done in the literature \cite{wu_integer_2020}, we reduce the learning rate to 1\% of the learning rate during pre-training and train for only 10\% of the epochs that the pre-trained full-precision model was trained for. Table \ref{tbl:smnist-qaft-overview} shows the results of QAFT after one epoch of training and 15 epochs of training (10\% of the 150 pre-training epochs).
It can be seen that QAT outperforms both PTQ and QAFT. This gap between QAT and QAFT may be negligible for larger precision (only 0.01 pp decrease for W8A8) but it becomes more significant for lower precision (0.43 pp decrease for W4A8\={A}8 and 0.36 pp for W2A8SSM8). We expect this trend to be even more pronounced for more complex tasks beyond sMNIST. 

\begin{table}[ht]
\centering
\caption{
Comparison of QAT, PTQ and QAFT (after 1 epoch and after full 15 epochs) on sMNIST.
The best-performing model for each quantization configuration is highlighted in \textbf{bold} (if it achieves at least 99\% test accuracy). The best overall model is \underline{underlined.}
The full-precision model achieves 99.65\% test accuracy on sMNIST.
}
\begin{tabular}{lcccc}
\toprule
            & QAT   & PTQ   & QAFT (1 epoch) & QAFT (15 epochs) \\
\midrule
\midrule
W8A8        & \textbf{99.54}                & 96.27 & 99.30 & 99.53 \\
W4A8SSM8    & \underline{\textbf{99.63}}    & 95.99 & 99.22 & 99.52 \\
W4A8\={A}8  &\textbf{ 99.26}                & 37.98 & 94.22 & 98.83 \\
W4A8        & 12.68                         &  9.74 &  9.74 & 10.93 \\
W2A8SSM8    & \textbf{99.56}                & 61.94 & 98.43 & 99.20 \\
W2A8\={A}8  & 80.76                         & 11.35 & 30.67 & 67.55 \\
W2A8        & 54.75                         & 11.35 & 17.60 & 35.84 \\
\bottomrule
\end{tabular}
\label{tbl:smnist-qaft-overview}
\end{table}

\end{document}